\documentclass{article}
\usepackage{ijcai17}

\usepackage{times}
\usepackage{epsfig}
\usepackage{graphicx}
\usepackage{amsmath}
\usepackage{amssymb}
\usepackage{mathsymb}
\usepackage{textcomp}
\usepackage{verbatim}
\usepackage{subfigure}
\usepackage{url}
\usepackage{graphicx}
\usepackage{multirow}
\usepackage[super]{nth}
\usepackage{colortbl}
\graphicspath{{./}{./Fig/}{./Figs/}{./Fig/eps/}{./Fig/pdf/}}

\usepackage{pbox}
\usepackage{algorithm2e}

\newcommand{\etal}{\textit{et al.}}
\newcommand{\eg}{\textit{e.g.,}}
\newcommand{\ie}{\textit{i.e.,}}

\title{Where to Focus: Deep Attention-based Spatially Recurrent Bilinear Networks for Fine-Grained Visual Recognition}
\author{Lin Wu$^{\ddag}$, Yang Wang$^{\dag}$\\
 $^{\ddag}$The University of Queensland, Australia\\
 $^{\dag}$The University of New South Wales, Kensington, Sydney, Australia\\
 lin.wu@uq.edu.au; wangy@cse.unsw.edu.au
}

\begin{document}

\maketitle

\begin{abstract}
Fine-grained visual recognition typically depends on modeling subtle difference from object parts. However, these parts often exhibit dramatic visual variations such as occlusions, viewpoints, and spatial transformations, making it hard to detect.  In this paper,  we present a novel attention-based model to automatically, selectively and accurately focus on critical object regions with higher importance against appearance variations.  Given an image, two different Convolutional Neural Networks (CNNs) are constructed, where the outputs of two CNNs are correlated through bilinear pooling to simultaneously focus on discriminative regions and extract relevant features. To capture spatial distributions among the local regions with visual attention, soft attention based spatial Long-Short Term Memory units (LSTMs) are incorporated to realize spatially recurrent yet visually selective over local input patterns. All the above intuitions equip our network with the following novel model: two-stream CNN layers, bilinear pooling layer, spatial recurrent layer with location attention are jointly trained via an end-to-end fashion to serve as the part detector and feature extractor, whereby relevant features are localized and extracted attentively. We show the significance of our network against two well-known visual recognition tasks: fine-grained image classification and person re-identification.
\end{abstract}

\section{Introduction}\label{sec:intro}

Fine-grained recognition such as identifying the \emph{species of birds, models of aircrafts, identities of persons}, is a challenging task since the realistic images between categories often have subtle visual difference, and easily overwhelmed by nuisance factors such as the poses, viewpoints or illuminations. Unlike general object recognition, fine-grained scenario can be improved by learning critical regions of the objects, to discriminate different subclasses meanwhile align the objects from the same class \cite{Part-stacked-CNN,DeepLAC,Weakly-fine,Mask-CNN}.

\begin{figure}[t]
\centering
\includegraphics[width=9cm,height=3cm]{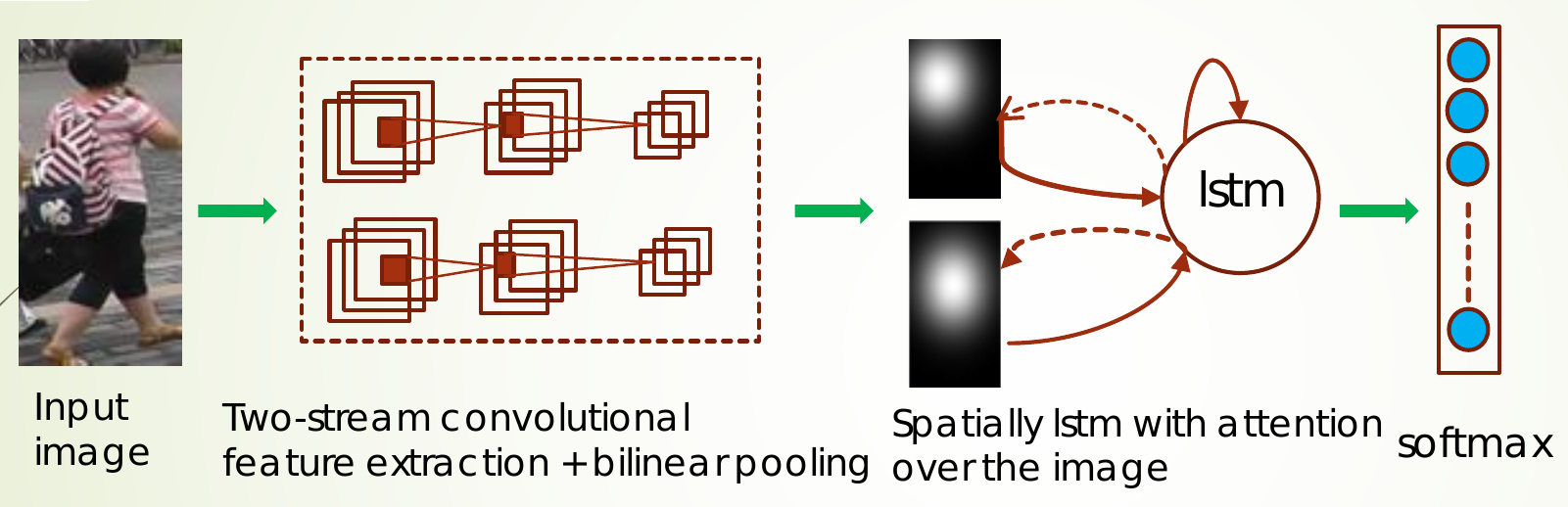}
\caption{The attention model with spatially recurrent bilinear features. The two-stream CNNs detect and extract features of regions, and spatially LSTMs focus selectively on distinct parts with visual attention. The white regions show what the mode is attending to and the brightness indicates the strength of focus. Best view in color.}\label{fig:introduction_fig}
\end{figure}
\subsection{Motivations}
The great efforts have been paid for fine-grained recognition. Motivated by the advances in training deep neural networks, the recent typical work \cite{BilinearCNNs,YangTIP2017,ACMMM15,Multiregion} has largely improved that by utilizing a principled bilinear CNNs to localize discriminative regions and model the appearances conditioned on them. Specifically, the feature extraction is performed based on the convolutional layers from two different CNN streams whose outputs are multiplied using outer product on each region (a.k.a bilinear pooling), on which sum-pooling over all regions is performed to derive global image descriptor. The resulting \emph{orderless} features are normalized, and fed into the softmax layer for classification. However, such spatial relationship is disposed in bilinear combination. Essentially, convolutional layers are using sliding filters, and their outputs, known as feature maps, which involve not only the strength of the responses, but also their spatial positions. This indicates that the matching of visual objects should follow its spatial constraints. For instance, the region containing the \emph{head of a bird} should be compared with \emph{head region} rather than \emph{feet}.

For fine-grained recognition, humans often abstract discriminative features of these objects and then compare the similarity among them to find the specific one, which can be repeated many times with relative spatial distributions (\eg multiple glimpses of each person on his/her hair, jacket, and pants). As indicated \cite{DynamicScene}, one of the most curious facets of the human visual system is ``attention". Rather than compress an entire image into static representation, ``attention" enables salient features to dynamically come to the forefront as needed. This is especially important when there are a lot of clutters in an image, and also in the case of visually similar objects that are difficult to be distinguished.
\subsection{Our approach}
Inspired by the observations above, we propose a soft attention model with spatial recurrence for fine-grained visual recognition, and show how our model dynamically pools convolutional features.
Specifically, we propose a flexible solution by simulating the process of human visual system and learning an end-to-end model from raw images to recurrently localize discriminative parts of visual objects in spatial manipulations. The network is designed to \emph{nontrivially} combine the strengths of bilinear CNNs and multi-dimensional recurrent neural networks (MDRNNs) \cite{MDRNN} to produce spatially expressive representations of feature interactions on critical object regions. Hence, the learned deep features help discriminate different subclasses (see Fig.\ref{fig:introduction_fig}). To this end, we implement a spatial variant of MDRNNs with long-short term memory units \cite{OfflineMDRNN}, \ie \textit{spatial LSTMs}, which naturally render bilinear pooled features spatially-context aware due to its capacity to capture long-range correlations. These recurrent connections are dynamically pooled via our soft attention mechanism to create flexible internal representations on focused object regions, yet robust to localized distortions along any combination of the input dimensions.\\
\textbf{Contributions} The major contributions are three-fold:
(1) We present a novel deep spatially recurrent model with visual attention. The proposed method is based on spatial LSTMs that benefits the bilinear features with spatial manipulations to focus on the most relevant regions but also to render those regions robust against potential transformations and local distortions;
(2) We delivered an interesting intuitions on our network from visualizing ``where'' and ``what'',  the attention focused on through stochastic back-propagation of classification errors in an end-to-end fashion;
(3) We show that using these features for visual recognition delivers better results compared with spatial pyramid pooling \cite{SPP-Net}. Our model is a  powerful mechanism to learn to attend at the right regions to extract relevant information for recognition, which outperforms the state-of-the-arts throughout the typical visual recognition tasks.

\vspace{-0.2cm}
\section{Related work}\label{sec:related}

\subsection{Fine-grained visual object recognition}

A number of effective fine-grained recognition methods have been developed in the literature \cite{Pose-CNN,Part-stacked-CNN,DeepLAC,Weakly-fine,LinYangarxiv,LinYang17hashingarxiv,LinTCYB2016,Linperson16,LinPR012017,LinPR2017,Mask-CNN,BilinearCNNs,SpatialTransformer,Part-R-CNN}. One pipeline is to align the objects to eliminate pose variations and the influence of camera positions, \eg \cite{Pose-CNN,DeepLAC}. Considering that the subtle difference between fine-grained images mostly resides in the unique properties of object parts, some approaches based on part-based representations \cite{Part-R-CNN,DeepLAC,Detection-part} use both bounding boxes of the body and part annotations during training to learn an accurate part localization model. The most related work to us are bilinear CNNs (B-CNNs) \cite{BilinearCNNs} and spatial transformer network \cite{SpatialTransformer}. B-CNNs introduce bilinear pooling upon the outputs of two different CNN streams in order to separate part detector and feature extractor. Nonetheless, the resulting bilinear features are orderless in which spatial relationship is disposed. In \cite{SpatialTransformer}, a module of spatial transformation is embedded into CNNs such that features from CNNs are invariant against a variety of spatial transformations. However, the region detection and feature extraction are not fully studied. By contrast, we enhance bilinear CNNs with visual attention in spatial recurrence which inherits the advantage of bilinear models and learns features not only robust against potential spatial transformations but also useful in salient region modeling.

\subsection{Long-short term memory with attention}

Attention models add a dimension of interoperability by capturing where the model is focusing its attention when performing a particular task. For example, a recent work of Xu \etal \cite{ShowAttendTell} used both soft attention and hard attention mechanism to generate image descriptions. Their model actually looks at the respective objects when generating their description. Building upon this work, Sharma \etal \cite{ActionAttention} developed recurrent soft attention based models for action recognition and analyzed where they focus their attention. However, that method is limited in learning representations in spatial constraints, which turn out to be crucial in visual recognition. In this paper, we propose recurrent attention model to generate location dependent features by learning to attend on spatial regions. To our best knowledge, our work is the first of realizing attention mechanism to fine-grained visual recognition.

\section{Deep spatially recurrent bilinear model with visual attention}

The attention model can be presented in a quadruple $\mathbb{M}=([g_A, g_B], \mathbf{B}, \mathbf{S}, \mathbf{C})$, where $g_A$ and $g_B$ are feature functions, $\mathbf{B}$ is bilinear pooling, $\mathbf{S}$ is a spatial recurrent function with LSTM units, and $\mathbf{C}$ is a classification function. In our architecture, each image is first processed by two separate CNNs ($g_A$ and $g_B$) to produce features at particular part locations. Then a bilinear pooling $\mathbf{B}$ allows pairwise correlations between feature channels and part detectors. After that, spatial LSTMs are used to model the spatial distribution of images with visual attention and produce hidden states as feature representation that can be fed into classification function $\mathbf{C}$. In the following sections, we will explain each of the components in greater details.

\subsection{Convolutional features}

We consider two CNNs to extract features to produce features for bilinear pooling. Specifically, we use CNNs pre-trained on the ImageNet dataset \cite{AlexNet}: \textbf{M-Net} \cite{M-Net} and \textbf{D-Net} \cite{VGG}, truncated at the convolutional layer including non-linearities as feature functions. The architectures of the two CNNs are shown in Table \ref{tab:convnet}. For notational simplicity, we refer to the complete CNNs as a function, $conv_5=g_A(\mathcal{I})$, $conv_5'=g_B(\mathcal{I})$ for the two CNNs, that take an image $\mathcal{I}$ as input and produces activations of the last convolution ($conv5$/$conv_{5\_4}$) as output.

\begin{table*}[t]\small
  \centering
  \caption{CNN architectures. Each architecture contains 5 convolutional layers (conv 1-5). In M-Net, the details of each convolutional layer are given in three sub-rows: the first specifies the number of convolution filters and their receptive field size as ``num $\times$ size $\times$ size''; the second indicates the convolution stride (``st.'') and spatial padding (``pad''); the third indicates if Local Response Normalisation (LRN)  is applied, and the max-pooling downsampling factor. In D-Net, each convolutional layer has additional $1\times 1$ convolution filters (\eg $conv_{1\_2}$), which can be seen as linear transformation of the input channels. The convolution stride is fixed to 1 pixel, and padding is 1 pixel for $3\times 3$ convolution layers. }  \label{tab:convnet}
  {
  \begin{tabular}{l|c|c|c|c|c}
  \hline
\hline
    Arch.  & conv1  &  conv2 & conv3 & conv4 & conv5 \\
  \hline\hline
M-Net & \pbox{5cm}{$96\times 7\times 7$ \\ st. 2, pad 0 \\LRN, $\times$2 pool} & \pbox{5cm}{$256\times 5\times 5$ \\ st. 2, pad 1 \\LRN, $\times$2 pool}  & \pbox{5cm}{$512\times 3\times 3$ \\ st. 1, pad 1 \\- }  & \pbox{5cm}{$512\times 3\times 3$ \\ st. 1, pad 1 \\ -} & \pbox{5cm}{$512\times 3\times 3$ \\ st. 1, pad 1 \\ -} \\
\hline
D-Net &\pbox{5cm}{$conv_{1\_1} (64\times 3\times 3)$ \\ $conv_{1\_2} (64\times 1\times 1)$ \\ $\times$ 2 pool} & \pbox{5cm}{$conv_{2\_1} (128\times 3\times 3)$ \\ $conv_{2\_2} (128\times 1\times 1)$ \\ $\times$ 2 pool} & \pbox{5cm}{$conv_{3\_1} (256\times 3\times 3)$ \\ $conv_{3\_2} (256\times 1\times 1)$ \\  $conv_{3\_3} (256\times 1\times 1)$ \\  $conv_{3\_4} (256\times 1\times 1)$} & \pbox{5cm}{$conv_{4\_1} (512\times 3\times 3)$ \\ $conv_{4\_2} (512\times 1\times 1)$ \\  $conv_{4\_3} (512\times 1\times 1)$ \\  $conv_{4\_4} (512\times 1\times 1)$} & \pbox{5cm}{$conv_{5\_1} (512\times 3\times 3)$ \\ $conv_{5\_2} (512\times 1\times 1)$ \\  $conv_{5\_3} (512\times 1\times 1)$ \\  $conv_{5\_4} (512\times 1\times 1)$} \\
\hline
  \hline
  \end{tabular}
  }
\end{table*}

\subsection{Bilinear pooling}

In CNNs, a feature function is defined as a mapping that takes an input image $\mathcal{I}$ at location $\mathcal{L}$ and outputs the feature of determined size $\mathcal{D}$, that is, $g: \mathcal{I}\times\mathcal{L} \rightarrow \mathcal{R}^{1 \times \mathcal{D}}$. Let $g_A \in \mathcal{R}^{a\times a \times D_A}$ and $g_B\in \mathcal{R}^{K\times K \times D_B}$ denote the feature outputs from the last convolution, where $K\times K$ is the feature dimension, $D_A$ and $D_B$ denote respective feature channels.
Feature outputs from the two feature extractors are combined at each location using the outer product, \ie \emph{bilinear pooling} operation of $g_A$ and $g_B$ at a location $l$:
\begin{equation}
\mathbf{B}(l,  \mathcal{I}, g_A, g_B)=g_A(l, \mathcal{I})^T g_B(l, \mathcal{I}).
\end{equation}
Thus, the bilinear form $\mathbf{B}\in \mathbb{R}^{K\times K \times \hat{D}}$ ($\hat{D}=D_A D_B$) allows the outputs of two feature streams to be conditioned on each other by considering all their pairwise interactions.
Specifically, in our architecture, the outer product of the deep features from two CNN streams are calculated for each spatial location, resulting in the quadratic number of feature maps.

Our intention here is to fuse two networks such that channel responses at the same position are put in correspondence. To motivate this, consider the case of recognizing a bird, if a filter in a CNN has responses to textures of some spatial location (head or wing), and the other network can recognize the location, and their combination then discriminates this bird species. To sequentially focus on different parts of the visual object and extract relevant information, bilinear features are filtered by location-dependent importance, which takes the expectation of the whole 2-D features, \ie $\mathbf{I}=E_{\mathbf{L}}  [\mathbf{B}]$ where $\mathbf{L}$ is a location matrix over $K\times K$ locations, which encodes the strength of focus (defined in Eq. \eqref{eq:location}).

\subsection{Spatial LSTM with soft attention mechanism}

\begin{figure}[t]
\centering
\includegraphics[width=5cm,height=2.5cm]{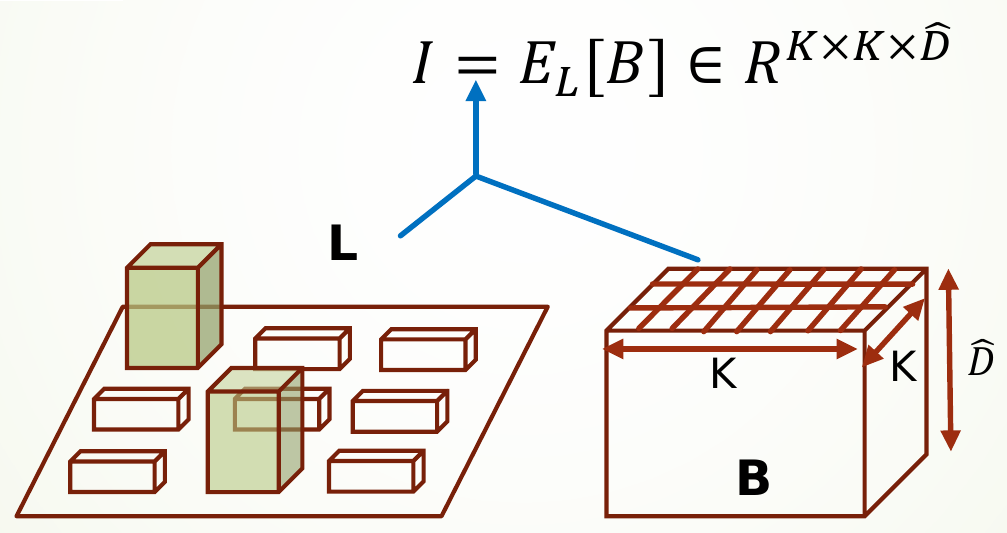}
\caption{The soft attention mechanism.}\label{fig:attention_location}
\end{figure}

The resulting bilinear features only allow feature interaction at every location on the spatial grid of last convolution outputs. The statistical correlations among grids should be captured in order to make the model flexible in local feature displacement and robust to potential spatial transformations. Meanwhile, different parts of an object should be assigned different weights from which more relevant informations can be retained. To model the distribution of locations, we employ spatial LSTM \cite{OfflineMDRNN}, which can be modified with location attention mechanism to achieve adaptive visual attention. For each location $(i,j)$ on a two-dimensional grid of $\mathbf{I}$, the operations performed by spatial LSTM are given by
\begin{equation}\label{eq:spatial_LSTM}\small
\begin{split}
&\mathbf{i}_{i,j} = \sigma( W_{xi} \mathbf{I}_{<i,j} + W_{hi}^r h_{i, j-1} + W_{hi}^l h_{i-1,j})\\
&\mathbf{f}_{i,j}^l = \sigma(W_{xf}^l \mathbf{I}_{<i,j} + W_{hf}^l h_{i-1,j})\\
&\mathbf{f}_{i,j}^r = \sigma(W_{xf}^r \mathbf{I}_{<i,j} + W_{hf}^r h_{i,j-1})\\
&\mathbf{o}_{i,j} = \sigma(W_{xo} \mathbf{I}_{<i,j} +W_{ho}^l h_{i-1,j} + W_{ho}^r h_{i,j-1})\\
&\mathbf{g}_{i,j} = \mbox{tanh}(W_{xc} \mathbf{I}_{<i,j} + W_{hc}^l h_{i-1,j} + W_{hc}^r h_{i,j-1})\\
&\mathbf{c}_{i,j} = \mathbf{g}_{i,j} \odot\mathbf{i}_{i,j} + \mathbf{c}_{i,j-1} \odot \mathbf{f}_{i,j}^r + \mathbf{c}_{i-1,j} \odot \mathbf{f}_{i,j}^l\\
&\mathbf{h}_{i,j}= \mbox{tanh}(\mathbf{c}_{i,j} \odot  \mathbf{o}_{i,j})
\end{split}
\end{equation}
where $\sigma$ is the sigmoid function, $\odot$ indicates an element-wise product. $W$ are the weights connecting the layers of the neurons. Let $\mathbf{I}_{i,j}$ be the feature at location $(i,j)$, and $\mathbf{I}_{<i,j}$ designate the set of locations $\mathbf{I}_{m,n}$ such at $m<i$ or $m=i$ and $n<j$. In our model, we limit $\mathbf{I}_{<i,j}$ to a smaller neighborhood surrounding the specific location, referred to \textit{casual neighborhood}. This is based on the assumption that each location is stationary to local displacement or shift invariance.
\vspace{-0.8cm}
\paragraph{Deterministic soft attention}
The feature cube of $\mathbf{I}$ is computed by multiplying location attention matrix over bilinear features $\mathbf{B}$: $\mathbf{I}=E_{\mathbf{L}}  [\mathbf{B}]$. This formulates a deterministic attention model by computing a soft attention weighted bilinear features. Specifically, the location attention is formulated into a location dependent matrix, $\mathbf{L}$, which is a softmax over $K\times K$ locations. The location softmax is defined as follows,
\begin{equation}\label{eq:location}\small
\mathbf{L}_{u,v}=P(\mathbf{L}=(u,v)| \mathbf{h}_{i,j})\\
=\frac{\exp (U^T_{u,v} \mathbf{h}_{i,j} )}{\sum_{u=1}^i \sum_{v=1}^j (U^T_{u,v} \mathbf{h}_{i,j}) },
\end{equation}
where $U_{u,v}$ are the weights mapping to the $(u,v)$ element of the location softmax, and $\mathbf{L}$ is a random variable which can take 1-of-$K^2$ values. This softmax can be thought of as the probability with which our model deems the corresponding region in the input frame is important. After calculating these probabilities, the soft attention mechanism \cite{NeuralMachine,ShowAttendTell} computes the expected values of the input by taking expectation over the feature slices at different regions (see Fig.\ref{fig:attention_location}):
\begin{equation}\small
\mathbf{I}=E_{P({\mathbf{L}_{u,v} | \mathbf{h}_{i,j}})}  [\mathbf{B}_{i,j}]=\sum_{u=1}^i \sum_{v=1}^j \mathbf{L}_{u,v}\mathbf{B}_{u,v}.
\end{equation}
where $\mathbf{B}$ is the bilinear pooled feature tube, and $\mathbf{B}_{u,v}$ is the ($u,v$) slice of the feature cube within the region $\mathbf{B}_{i,j}$. This corresponds to feeding in a soft weighted feature cube into the system, and the whole model is smooth and differentiable under the deterministic attention. Thus, learning end-to-end is trivial by using standard back-propagation.

\subsection{Loss function and attention penalty}

In the training of our model, we use cross-entropy loss coupled with the doubly stochastic penalty regularization \cite{ShowAttendTell}, which encourages the model to pay equal attention to every part of the image. We impose an additional constraint over the location softmax, so that $\sum \mathbf{L}_{u,v}\approx \tau$ where $\tau\ge K^2/\hat{D}$. Finally, the loss function is defined as follows:
\begin{equation}\small
Loss=\sum_{i=1}^C y_i \log \hat{y}_i + \lambda \sum_{u,v} (\tau-\mathbf{L}_{u,v})^2 +\gamma \sum_i\sum_j \Theta^2_{i,j},
\end{equation}
where $y$ is the one hot label vector, $\hat{y}$ is the vector of class probabilities, which is computed via Eq.\eqref{eq:class_vector}, $C$ is the number of output classes, $\lambda$ is the attention penalty coefficient, $\gamma$ is the weight decay coefficient, and $\Theta$ represents all the model parameters. The computation of $\hat{y}$ is
\begin{equation}\label{eq:class_vector}
P(\hat{y}=j|x)=\frac{e^{x \cdot w_j}}{\sum_{c=1}^C e^{x \cdot w_c}}
\end{equation}
where $x$ is the feature vector from the network, and $P(\hat{y}=j|x)$ is to predict the probability for the $j$-th class given $x$ over a combination of $C$ linear functions.
The gradients of classification, recurrent layer of LSTM units, bilinear layer, and two-stream CNNs with convolution, pooling, and non-linear activations can be computed using the chain rule. We use the following initialization strategy \cite{ShowAttendTell} for the cell states, and hidden states of spatial LSTM for faster convergence:
\begin{equation}\small
\mathbf{c}_0=f_{init,c}\left(\frac{1}{K^2} \sum_{u}^K\sum_{v}^KB_{u,v}\right), \mathbf{h}_0=f_{init,h}\left(\frac{1}{K^2} \sum_{u}^K\sum_{v}^KB_{u,v}\right)
\end{equation}
where $f_{init,c}$ and $f_{init,h}$ are two multi-layer perceptions and these values are used to calculate the first location matrix $\mathbf{L}$ which determines the initial input  of $\mathbf{I}$. Details about architecture and hyper-parameters are given in Section \ref{ssec:implement}.

\section{Experiments}\label{sec:exp}

\begin{figure}[t]
\includegraphics[width=4cm,height=2cm]{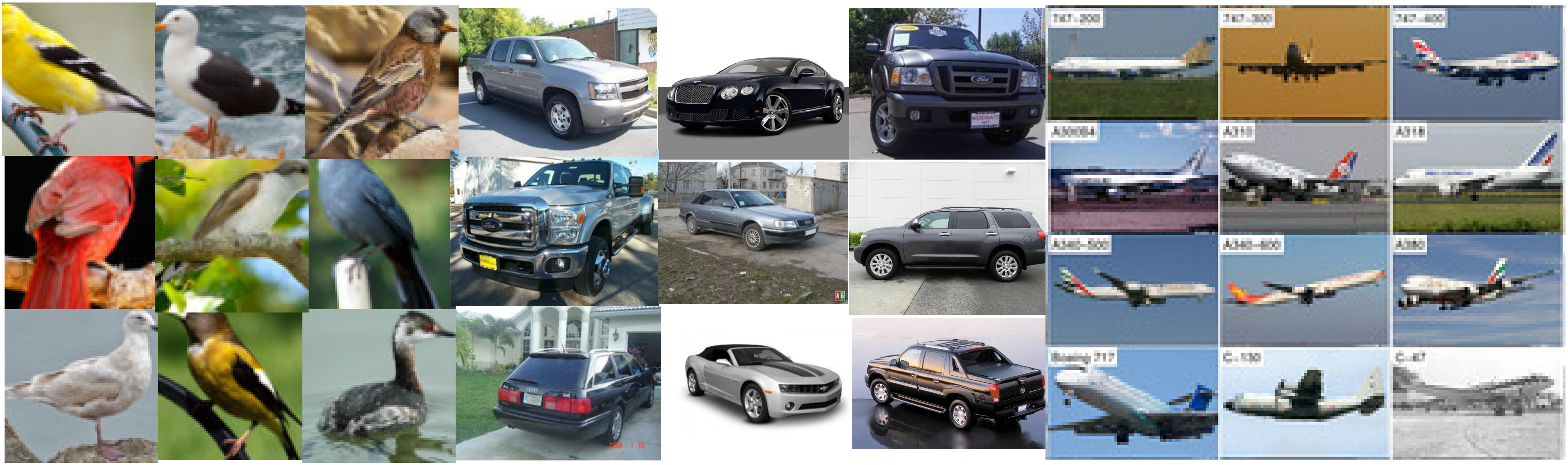}
\includegraphics[width=4cm,height=2.2cm]{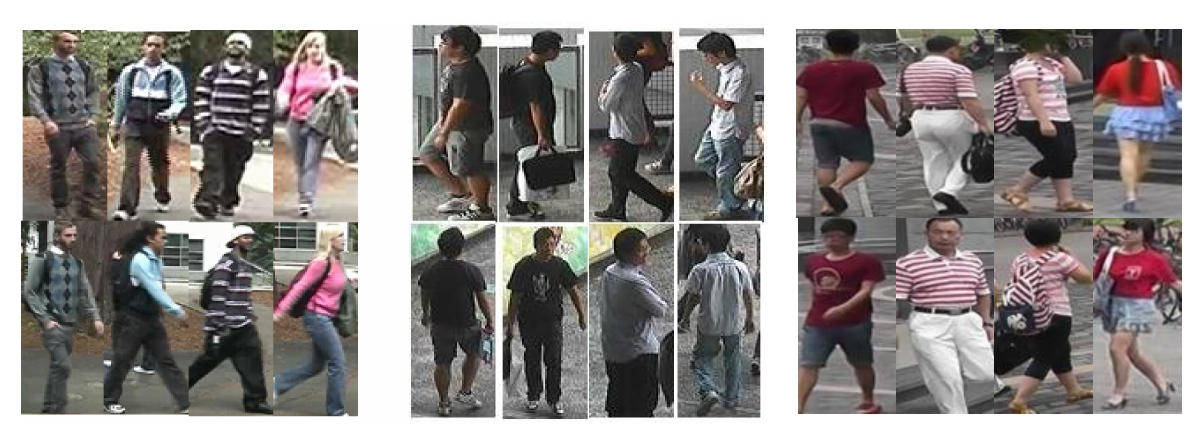}
\caption{Examples from birds dataset (left), cars dataset (middle), aircraft dataset (right), and examples from person re-identification dataset: VIPeR (left), CUHK03 dataset (middle), and Market1501 (right).}\label{fig:datasets}
\end{figure}

To evaluate the effectiveness of the proposed method, we conduct experiments by comparison to a variety of baselines and state-of-the-arts on two applications: fine-grained image classification and person re-identification.

\paragraph{Why on fine-grained image categorization and person re-identification?}
Fine-grained recognition tasks refer to test the property of the proposed method in localizing object parts and modeling the appearance conditioned on detected locations while being robust against a range of spatial transformations. Person re-identification shares much similar to fine-grained categorizations where the matching process often resorts to the analysis of texture details and body parts to be localized. Also, the spatial distribution among distinguished parts is a helpful prior in recognizing identities.

\subsection{Baselines}
The following baselines are used in our experiments:
(1) CNN with fully-connected layers (FC-CNN):
The input image is resized to $224\times 224$ and mean-subtracted before propagating it through the CNN. For fine-tuning, we replace the 1000-way classification layer trained on the ImageNet with a $k$-way softmax layer where $k$ is the number of classes in target dataset. The layer before the softmax layer is used to extract features.
(2) Fisher vectors with CNN features (FV-CNN):
Following \cite{FilterBanks}, we construct a descriptor using FV pooling of CNN filter bank responses with 64 GMM components. FV is computed on the output of the last convolution layer. Following \cite{BilinearCNNs}, the input images are resized to $448\times 448$ and pool features in a single-scale.
(3) Fisher vectors with SIFT features (FV-SIFT):
It is implemented by using dense SIFT features over 14 dense overlapping $32\times 32$ pixels regions with a step stride of 16 pixels in both direction. The features are PCA projected before learning a GMM with 256 components.
(4) Bilinear CNN classification model (B-CNN) \cite{BilinearCNNs}: This method is to perform bilinear pooling on the output features from two CNN streams. Then, orderless sum-pooling is employed to aggregate the bilinear features across the image. We use the model initialized with a D-Net and a M-Net (B-CNN [D,M]). The input images are resized to $448\times 448$ and features are extracted using two networks before bilinear combination, sum-pooling and normalization. The D-Net produces output $28\times 28$ while M-Net has $27\times 27$. Thus, a down-sampling is conducted by dropping a row and a column from D-Net outputs.
(5) Bilinear CNN model with spatial pyramid pooling \cite{SPP-Net} (B-CNN+SPP):
To have fair comparison, we perform a 2-level pyramid \cite{SPP-Net}: $2\times 2$ and $1\times 1$ subdivisions.
(6) The proposed method: Identical to B-CNN, features are extracted using two CNNs with outputs from the last convolutional layer ($conv5$+relu for M-Net and $conv_{5_4}$+relu for D-Net), followed by bilinear pooling, spatial recurrence with visual attention, and flattening.

\begin{table}[t]\small
  \centering
  \caption{Fine-grained categorization results. We report per-image accuracy on three datasets: CUB-200-2011, without bounding-boxes on birds body part, aircrafts and cars.}  \label{tab:fine_grained}
  {
  \begin{tabular}{l|c|c|c}
  \hline
\hline
    Method  & Birds  &  Aircrafts & Cars  \\
  \hline\hline
FV-SIFT &18.8 & 61.0& 59.2\\
FC-CNN [M] & 58.8 & 57.3& 58.6\\
FC-CNN [D] & 70.4& 74.1& 79.8\\
FV-CNN [M] & 64.1& 70.1& 77.2\\
FV-CNN [D] & 74.7 & 77.6& 85.7\\
B-CNN [D,M] & 84.1& 83.9& 91.3\\
B-CNN + SPP [D,M] & 86.9 & 86.7 & 92.5\\
\hline
Krause \etal  & 82.0 & -& 92.6 \\
Part-based R-CNN & 73.9 & -&-\\
Pose-normalized CNN  & 75.7 & -&-\\
Spatial transformer  & 84.1 & - & -\\
Chai \etal  & -& 72.5 & 78.0\\
Gosselin \etal&- & 80.7 & 82.7\\
\hline
Ours [D,M] &  \textbf{89.7} & 88.4 & \textbf{93.4}\\
  \hline
  \end{tabular}
  }
\end{table}

\subsection{Implementations}\label{ssec:implement}
In all of our experiments, model architecture and hyper-parameters are set using cross-validation. In particular, we train a 2-layer spatial LSTM model for all datasets, where the dimensionality of the hidden state and cell state are set to 512. The attention penalty coefficient $\lambda$ is set to be 1, weight decay $\gamma=10^{-5}$, and use dropout rate of 0.5 at all non-recurrent connections.
In fact, the training of proposed model is fine-tuning the components of of pre-trained two CNNs, and spatial recurrence. To this end, we add a $k$-way softmax layer. We adopt the two-step training procedure \cite{Pose-CNN} where we first train the last layer using logistic regression, followed by a fine-tuning the entire model using back-propagation for a number of epochs at a small learning rate ($\eta=0.001$).
Once the fine-tuning is done, training and validation sets are combined to train one-vs-all linear SVMs on the extracted features.
In experiments, we employ two kinds of data augmentation: flipping and shifting. For flipping, we flip each sample horizontally to allow the model observe mirror images of the original images during training. For shifting, we shift each image by 5 pixels to the left, 5 pixels to the right, and then further shift it by 3 pixels to the top, and 3 pixels to the bottom. This procedure makes the model more robust to slight shifting of an object in an image. The shifting was done without padding the borders of the images.

\begin{table*}[t]\small
  \caption{Rank-1, -5, -10, -20 recognition rate of different methods on the VIPeR, CUHK03 dataset, and mAP on Market1501. }\label{tab:cmc_value}
  {
  \begin{tabular}{l|c|c|c|c|c|c|c|c|c|c}
  \hline
\hline
& \multicolumn{4}{c|}{VIPeR} & \multicolumn{4}{c|}{CUHK03} & \multicolumn{2}{c}{Market1501}\\
\cline{2-11}
    Method  & $R=1$  &  $R=5$ & $R=10$  & $R=20$ & $R=1$  &  $R=5$ & $R=10$  & $R=20$  & $R=1$  &  mAP\\
  \hline\hline
   JointRe-id \cite{JointRe-id} & 34.80 & 63.32  & 74.79 & 82.45  & $54.74$ & 86.42 & 91.50 & 97.31 &-&-\\
   SDALF \cite{Farenzena2010Person} & 19.87 & 38.89 & 49.37 & 65.73 & $5.60$ & 23.45 &36.09 & 51.96&  20.53 & 8.20\\
  SalMatch \cite{Zhao2013SalMatch} & 30.16 & 52.00 & 62.50 & 75.60 &-&-&-&-  & -&-\\
  PatchStructure \cite{Correspondence} & 34.80 & 68.70 & 82.30 & 91.80 &-&-&-&-  &-&-\\
  SCSP \cite{SimilaritySpatial} & 53.54 & 82.59 &  91.49 &  96.65 &-&-&-&-  & 51.90 & 26.35\\
 LocalMetric \cite{LocalMetric} & 42.30 & 70.99 & 85.23 & 94.25 &-&-&-&-  &-&-\\
 FPNN \cite{FPNN} &-&-&-&- & $20.65$ & 51.32 & 68.74 & 83.06 &-&-\\
 Multi-region \cite{Multiregion} &-&-&- &-& 63.87 & 89.25 & 94.33 & 97.05 & 45.58 &-\\
\hline
   Ours [D,M]&  $\mathbf{56.11}$  & $\mathbf{87.29}$ & $\mathbf{93.66}$ &  $\mathbf{96.97}$ & $\mathbf{65.23}$ &  $\mathbf{90.95}$ & $\mathbf{95.73}$ & $\mathbf{98.52}$ & $\mathbf{54.67}$& $\mathbf{34.33}$\\
  \hline
  \end{tabular}
  }
\end{table*}

\subsection{Results on fine-grained image classification}

We conduct experiments on three fine-grained datasets: bird species \cite{Birds}, aircrafts \cite{Aircraft}, and cars \cite{Cars}. Examples selected from the three datasets are shown in Fig.\ref{fig:datasets}.

\subsubsection{Birds species classification}
The CUB-200-2011 dataset \cite{Birds} contains 11,788 images of 200 bird species. All methods are evaluated in a protocol where the object bounding-boxes are not provided in both training and testing phase.
The comparison results without bounding boxes are shown in Table \ref{tab:fine_grained}. We can see that FV-CNN[D] 74.7\% and FV-CNN[M] 64.1\% achieves better results than FC-CNN [D] 70.4\% and FC-CNN [M] 58.8\%. This is mainly because FV-CNN pools local features densely within the described regions, and  therefore more apt at describing local patch textures. Our method achieves the best results compared with B-CNN and B-CNN+SPP in all corresponding variants.
More recent results are reported by Krause \etal \cite{KrauseCVPR15} where 82\% accuracy is achieved by leveraging more accurate CNN models to train part detectors in a weakly supervised manner. Part based R-CNN \cite{Part-R-CNN} and pose-normalized CNN \cite{Pose-CNN} also perform well on this dataset with accuracy of 73.9\% and 75.7\%, respectively. However, the two methods are performing a two-step procedure on part detection and CNN based classifier. A competing accuracy of 84.1\% is achieved by spatial transformer networks \cite{SpatialTransformer} while this method only models the spatial transformation locally.

\subsubsection{ Aircraft classification}

The Fine-Grained Visual Classification of Aircraft (FGVC-Aircraft) dataset \cite{Aircraft}  consists of 10,000 images of 100 aircraft variants. The task involves discriminating variants such as Boeing 737-300 from Boeing 737-400, and thus the difference are very subtle, where sometimes one may be able to distinguish them by counting the number of windows in the model. In this dataset, airplanes tend to occupy a large portion of the whole image and appear in a relatively clear background.
Comparison results are reported in Table \ref{tab:fine_grained}, from which it can be seen that the results of trends are similar to those in birds dataset. It is notable that FV-SIFT performs remarkably better (61.0\%) and outperforms FC-CNN [M] (57.3\%).
In comparison to state-of-the-art approaches, the two best performing methods \cite{GosselinPRLetter} and \cite{Symbiotic} achieve 80.7\% and 72.5\%, respectively. Our method outperforms these approaches by a significant margin. It indicates that spatial pooling is vital to image categorization due to its robustness to local feature displacement.

\subsubsection{Car model classification}

The cars dataset \cite{Cars} contains 16,185 images of 196 classes. The data is split into 8,144 training images and 8,041 testing images, where each class has been split roughly in a 50-50 split. Classes are typically at the level of Make, Model, Year, \eg 2012 Tesla Model S or 2012 BMW M3 coupe. Cars in the dataset are smaller and appear in a more cluttered background, and thus, challenging object and part localization. Once again the proposed method consistently outperforms all baselines with [D, D] model achieving 93.4\% accuracy. Krause \etal \cite{KrauseCVPR15} achieves 92.6\%, and methods of \cite{GosselinPRLetter} and \cite{Symbiotic} achieve 82.7\% and 78.0\%. Our spatial bilinear model has a clear advantage over these models by attentively selecting features from regions for matching.


\subsection{Results on person re-identification}

We perform experiments on three benchmarks: VIPeR  \cite{Gray2007Evaluating}, and CUHK03 \cite{FPNN}. The \textbf{VIPeR} data set contains $632$ individuals taken from two cameras with arbitrary viewpoints and varying illumination conditions. The 632 person's images are randomly divided into two equal halves, one for training and the other for testing. The \textbf{CUHK03} data set includes 13,164 images of 1360 pedestrians. This dataset provides both manually labeled and detected pedestrian bounding boxes, and we report results on labeled data set. The dataset is randomly partitioned into training, validation, and test with 1160, 100, and 100 identities, respectively. The \textbf{Market-1501} data set contains 32,643 fully annotated boxes of 1501 pedestrians, making it the largest person re-id dataset to date. Each identity is captured by at most six cameras. The dataset is randomly divided into training and testing sets, containing 750 and 751  identities, respectively. The evaluation metric we use is Cumulative Matching Characteristic (CMC). This evaluation is performed ten times, and the average results are reported.

\subsubsection{Comparison to state-of-the-art approaches}

Comparative experiments with state-of-the-art methods are conducted, and results are reported in Table \ref{tab:cmc_value}. It can be seen that our approach outperforms all competitors consistently on the two benchmarks on rank-1 recognition accuracy. Compared with some approaches that consider pre-defined spatial distribution among body parts to improve matching such as SCSP \cite{SimilaritySpatial}, PatchStructure \cite{Correspondence}, SDALF \cite{Farenzena2010Person}, and SalMatch \cite{Zhao2013SalMatch}, our model is more beneficial to person re-identification by jointly performing feature extraction and spatial manipulation.  Compared with deep learning approaches with computation on local patch region difference \ie JointRe-id \cite{JointRe-id}, FPNN \cite{FPNN}, and LocalMetric \cite{LocalMetric}, our method learns features from critical parts, which helps discriminate different persons with subtle differences. Also, our method achieves performance gain over multi-region based bilinear models \cite{Multiregion} which manually partition body parts on which bilinear features are computed, whereas the proposed network can localize distinct patches with spatial attention and select features automatically for matching.
\vspace{-0.4cm}
\paragraph{Learning to attend}

\begin{figure}[t]
\centering
\includegraphics[width=4cm,height=2.2cm]{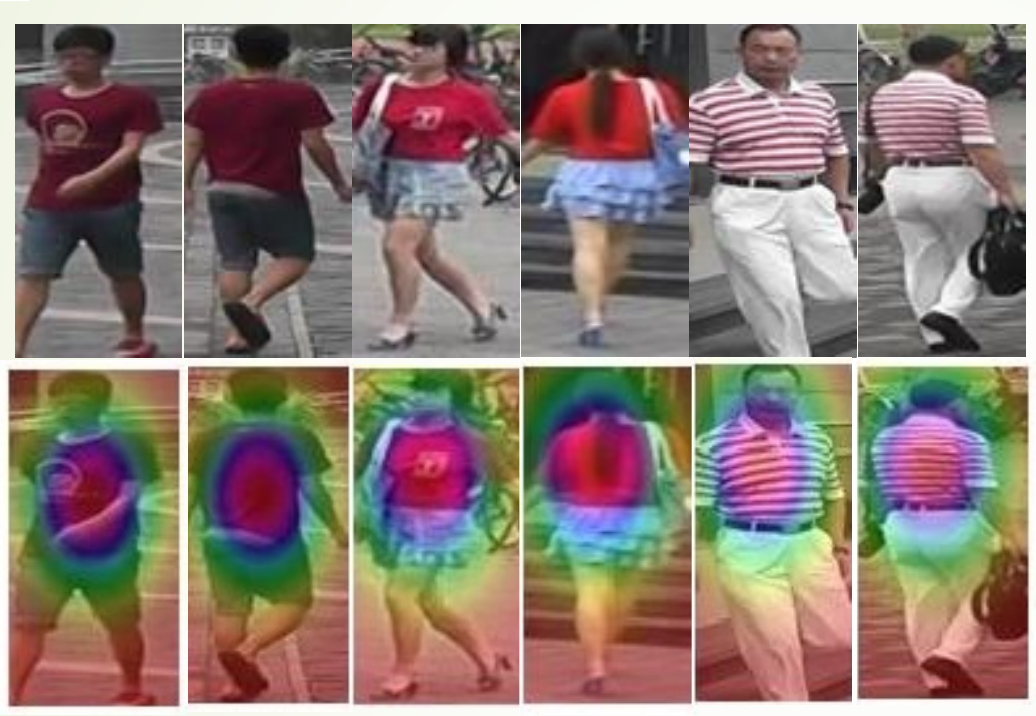}
\caption{Attending to distinct regions. Best view in color.}\label{fig:attend_results}
\end{figure}

Visualizing the attention learned by the model allows us to interpret the output of the model. In this sense, our model is more flexible by attending salient regions. The input to two CNNs is resized to $448\times 448$, and with five convolution with max pooling layers and bilinear pooling, we get an output dimension of $28\times 28$. In order to visualize the attention weights for the soft model, we upsample the weights by a factor of $2^5=32$ and apply a Gaussian filter emulate the large receptive field size. Fig. \ref{fig:attend_results} shows the model learns alignments that agree very strongly with human intuition.

\section{Conclusion}\label{sec:con}

In this paper, we present a deep recurrent soft attention based model for fine grained visual recognition and analyzed where they focus their attention. The proposed model tends to recognize important elements in visual objects that have subtle appearance difference, and thus achieve performance improvement. The impressive soft attention models are found to be computationally expensive since they require all the features to perform dynamic pooling. In future, we plan to explore some hard attention solution to sample locations over input image, as well as kernel approximation to bilinear features and multi-view features \cite{YangTIP2015,ijcai16,LinIVC2017,LinYangneuro,LinYangMTAP,ACMMM13,WangTNNLS,ArxivYangWang,Arxiv17YangWang,KAISYang16,WangINS,ACMMM14,ACMSIGIR15,PAKDD14,PAKDD142}, which can reduce the computational cost of our model.

\bibliographystyle{named}\small
\bibliography{ijcai17}

\end{document}